\documentclass[journal]{IEEEtran}
\usepackage{todonotes}
\usepackage{url}
\usepackage[normalem]{ulem}
\usepackage{colortbl}
\usepackage[capitalize]{cleveref}
\begin{document}
\title{Distributed AI in Zero-touch Provisioning for Edge Networks: Challenges and Research Directions}
\author{Abhishek~Hazra,~\IEEEmembership{Member,~IEEE,}
        Andrea~Morichetta,~\IEEEmembership{Member,~IEEE,} 
        Ilir~Murturi,~\IEEEmembership{Member,~IEEE,} 
        Lauri~Lovén,~\IEEEmembership{Senior~Member,~IEEE,}
        Chinmaya~Kumar~Dehury,~\IEEEmembership{Member,~IEEE,}
        Víctor~Casamayor~Pujol,~\IEEEmembership{Member,~IEEE,}
        Praveen~Kumar~Donta,~\IEEEmembership{Senior~Member,~IEEE,}
        Schahram~Dustdar~\IEEEmembership{Fellow,~IEEE}  
        
\thanks{P.K. Donta is the corresponding author}
    \thanks{A.~Hazra is with the Department of Computer Science and Engineering at the Indian Institute of Information Technology Sri City, India and Communications \& Networks Lab, Department of Electrical and Computer Engineering, National University of Singapore, Singapore 119260.}

\thanks{A. Morichetta, I. Murturi, V. C. Pujol, P. K. Donta, and S. Dustdar are with Distributed Systems Group, TU Wien, Vienna 1040, Austria.}
\thanks{L. Lovén is with Center for Ubiquitous Computing, University of Oulu 90014, Finland.}
\thanks{C. K. Dehury is with Institute of Computer Science, University of Tartu, Tartu 51009, Estonia.}
}

\markboth{IEEE Computer,~Vol.~XX, No.~X, November~2023}
{Regular issue}

\maketitle

\begin{abstract}
Zero-touch network is anticipated to inaugurate the generation of intelligent and highly flexible resource provisioning strategies where multiple service providers collaboratively offer computation and storage resources. This transformation presents substantial challenges to network administration and service providers regarding sustainability and scalability. This article combines Distributed Artificial Intelligence (DAI) with Zero-touch Provisioning (ZTP) for edge networks. This combination helps to manage network devices seamlessly and intelligently by minimizing human intervention. In addition, several advantages are also highlighted that come with incorporating Distributed AI into ZTP in the context of edge networks. Further, we draw potential research directions to foster novel studies in this field and overcome the current limitations.  
\end{abstract}

\begin{IEEEkeywords}
Distributed Artificial Intelligence; Edge Networks; Edge Resource Federation; Internet of Things; Zero-touch Provisioning
\end{IEEEkeywords}

\IEEEpeerreviewmaketitle

\section{Introduction}
The Internet of Things (IoT) has witnessed a rapid rise in recent years. Connected devices have integrated with various areas of our societies, such as electric grids, transportation, and industries~\cite{kokkonen2022autonomy}. This paradigm change leads to rethinking the processes related to IoT systems. Indeed, IoT systems that deal with people, connected devices, and data produced in their respective environments have to guarantee seamless management and integration of all these actors. The primary driving force in fostering this digital transformation is, thus, the integration of multiple functional systems to provide faster, steadier, more cost-effective, and overall better services. In particular, there's a need for automating resource provisioning in complex and broad scenarios like the device-edge-cloud computing continuum. Currently, most technologies, especially cloud-based services focus on centralized strategies ~\cite{gallego2021machine}. However, centralized approaches cannot work in these broad scenarios. In this regard, Zero-touch Provisioning (ZTP) represents an appealing direction. This class of approaches aims at providing methods to seamlessly and automatically manage the devices in a network, adapting to changes without requiring direct human intervention~\cite{gallego2021machine}. Such approaches are primarily implemented with Machine Learning (ML) methods.

The challenge is how to let AI approaches govern such complex systems \cite{gill2022ai}. Research on Distributed Artificial Intelligence (DAI) goes in this direction. DAI has seen waves of popularity over the decades~\cite{Chaibdraa2004TrendsID}. With AI research focus recently shifting towards decentralized and widespread systems, DAI is now studied with renewed interest. In particular, when managing networks with limited or no human contact, it is essential to separate knowledge and learning mechanisms over the infrastructure. This separation optimizes the system, makes it work at scale, and preserves privacy. DAI methods can be crucial in providing this separation, ensuring that individual nodes in the network can compute with local data only~\cite{augca2022survey}. Moreover, DAI methods can allow more accurate and robust prediction models by combining knowledge from many data sources separated by computational limits, administrative boundaries, or privacy restrictions. Furthermore, distributing AI minimizes resources or costs needed in moving the information. 

Still, there are open issues on letting multiple distributed computing agents communicate and produce effective solutions in real, complex and heterogeneous scenarios such as ZTP~\cite{chafika2021distributed}. The limited computational capabilities of devices in edge networks need novel methods to learn where to compute, how and when to distribute the data, how to guarantee an optimal model management to keep the performance adequate, and how to be security-aware.

\begin{figure}[!t]
    \centering
    \includegraphics[width=1.0\linewidth]{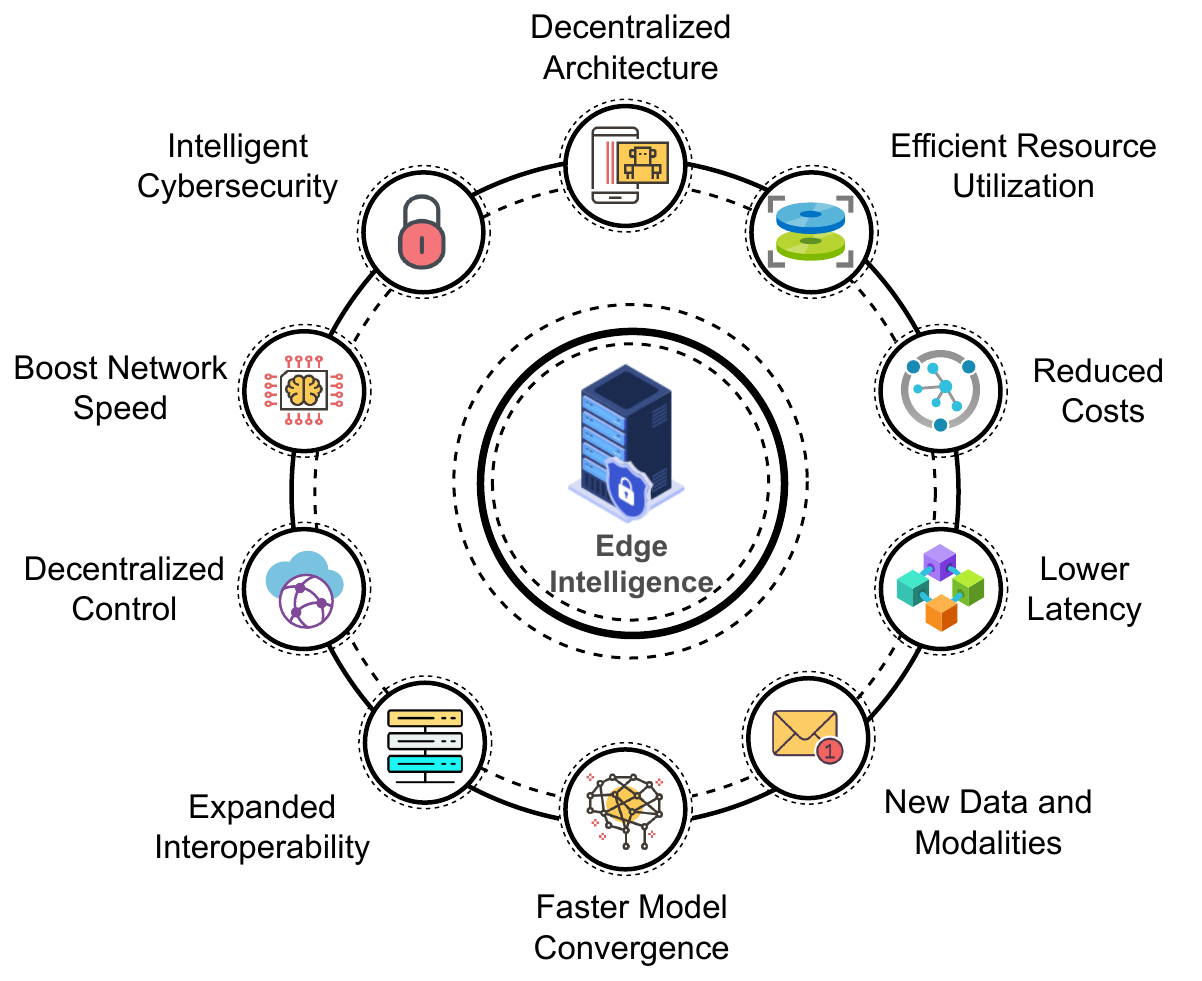}
    \caption{Benefits of zero-touch provisioning in the computing continuum.}
    \label{fig:zero-touch-benefits}
\end{figure}

The integration of DAI for Zero-touch provisioning at edge networks has intriguing implications, especially considering the advances proposed by 5G and 6G technologies~\cite{Chergui2021ZTouchAI}. In more detail, ZTP can foster several relevant advantages~\cite{benzaid2020ai}, as highlighted in Fig.~\ref{fig:zero-touch-benefits}. In summary, automating the logical setup of the network reduces the effort that IT teams have to put into the configuration phase. Most of the remaining work involves cabling and booting devices. Furthermore, when dealing with large and widely distributed networks,  the autonomy provided by ZTP reduces the time needed to operationalize the networks. Moreover, ZTP leads to less complex and more effective management at run-time, reducing the probability of human error and enabling faster updates.
Overall, it provides the ability to exploit large-scale computation and efficiently utilize spatially distributed computing resources in a decentralized manner with low-operating costs, low latency, faster model convergence, and decentralized control~\cite{Carrozzo2020AIdrivenZO}.
On the other hand, software-defined management is not risk-free. Misconfigurations caused by automated processes may be challenging to detect, leading to complex error detection. Furthermore, managing a network on a high abstraction level means exchanging data and potentially sensitive information, introducing security threats. Consequently, the ML methods must emphasize distributed and privacy-preserving intelligence~\cite{dustdar2022distributed}.
Finally, harmonically managing independent or partially correlated agents that make decisions in complex, heterogeneous systems requires studying new ways of coordination. 

To the best of our knowledge, this article is the first work targeting DAI in Zero-touch Provisioning for the computing continuum. The key objective is to introduce a novel edge computing architecture that combines DAI and ZTP into one platform and offers better services to the users. The major contributions of this article are as follows.
\begin{itemize}
    \item Design a ZTP-enabled edge computing architecture to support intelligent service provisioning while enhancing computation, communication, and storage functionalities to the users.
    \item We aim to highlight the challenges that come with DAI, ZTP, and their combination in the context of edge networks.
    \item We also discuss the network and service management challenges while offering computation and resource management solutions with ZTP in edge networks.
    \item  Finally, we introduce potential research directions that can foster novel studies in this field and overcome the current limitations. 
\end{itemize}
The remaining sections are structured as follows. \textit{Section}~\ref{section2} discusses the advantages of DAI in ZTP-enabled edge networks, where we introduce a novel edge computing architecture in the computing continuum. \textit{Section}~\ref{section3} highlights several potential challenges that still need to be addressed for the proper deployment of ZTP in edge networks. Finally, we conclude our discussion in \textit{Section}~\ref{section4}. 
\begin{figure*}[!t]
    \centering
    \includegraphics[width=1\linewidth]{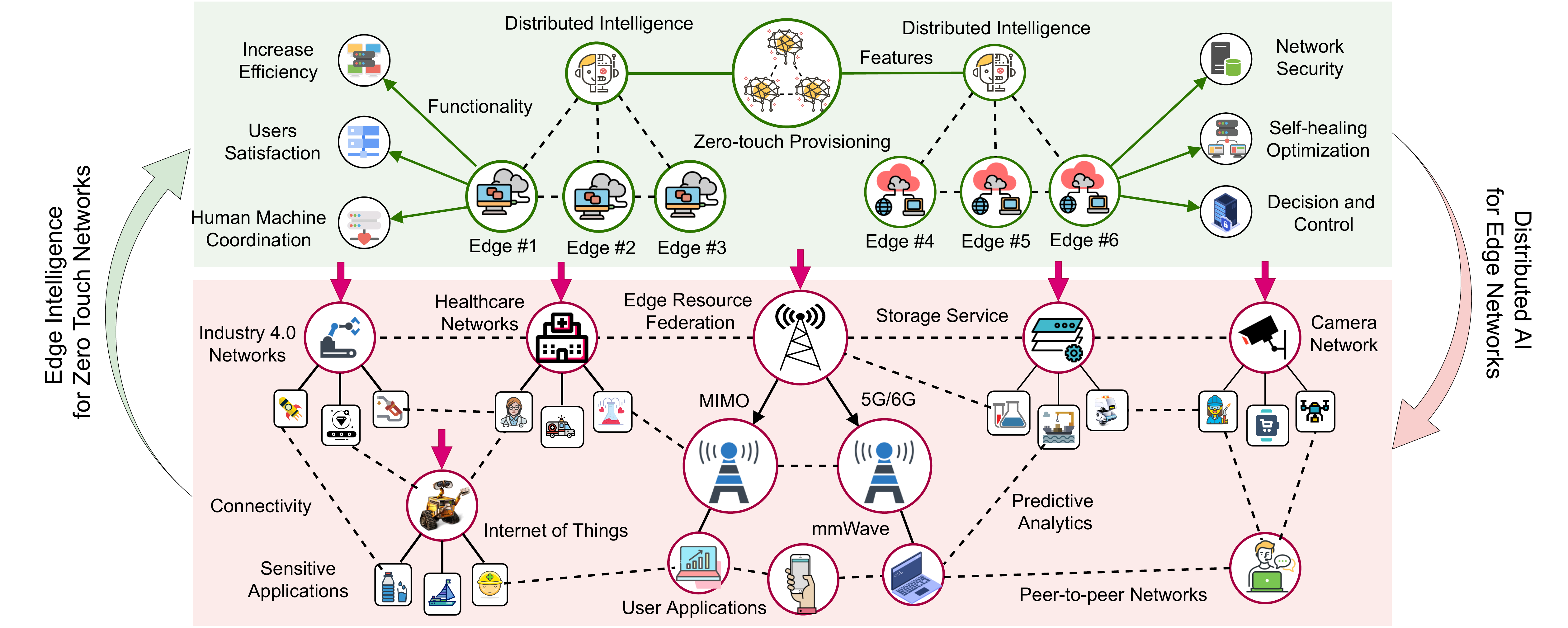}
    \caption{\textcolor{blue}{Zero-touch network in the computing continuum.}}
    \label{fig12}
\end{figure*}

\section{Distributed AI in ZTP for Edge Networks}\label{section2}
Traditional edge computing brings cloud services to the network's edge. However, edge computing, while offering computation, communication, storage, and resource solutions, has several evolving challenges related to end-to-end service and network management. To address such ripening challenges, we propose a new collaborative service management model, combining DAI, edge resource federation, and ZTP concepts. The edge-enabled ZTP framework is a trusted collection of services and related resources that intelligently integrates the infrastructures of distributed computing continuum service providers and ultra-reliable communication technologies to achieve low latency, scalability, and cost-efficient edge data transmission and processing, as shown in Fig.~\ref{fig12}. This framework consists of two major units:
{\color{black}
\begin{itemize}
    \item \textbf{Edge Intelligence for Zero Touch Networks:} Edge intelligence plays a crucial role in realizing the concept of Zero Touch Networks. Fundamentally, this technology facilitates data processing at a local level, granting edge devices the ability to independently assess and respond to data without relying on centralized decision-making. The use of distributed decision-making processes reduces latency, optimises network resources, and guarantees real-time responsiveness, as illustrated in Fig. 2. Edge intelligence enables the deployment of machine learning models at the edge, facilitating the implementation of predictive maintenance, anomaly detection, and dynamic load balancing. These features enable networks to function effectively, reduce the need for human involvement, and smoothly adjust to dynamic circumstances.
    \item \textbf{Distributed AI for Edge Networks:} DAI facilitates the deployment of AI capabilities to the periphery of network infrastructures. Edge devices equipped with AI models can make real-time decisions, process data locally, and function independently. Recent advances in DAI exhibits high scalability, making it well-suited to networks that experience an increasing quantity of devices and services. AI models have the capability to be tailored to fulfil specified criteria, hence guaranteeing self-governance and protection of edge operations. AI deployment at the edge of networks has several advantages, including improved efficiency, decreased reliance on centralized control, and the fulfilment of Zero Touch Network and Service Management objectives.    
\end{itemize}

}
In summary, edge computing and ZTP technology can present new business opportunities to network operators, heterogeneous IoT users, and cloud service providers. Further combining DAI and edge resource federation in the ZTP networks, customers can experience rapid data accessibility, seamless network coverage, interoperable data migration, and innovative services, which will eventually help to enhance user happiness~\cite{luque2022toward}. The key techniques of edge-enabled ZTP networks are discussed below.

\begin{table*}[t]
\label{table1}
\caption{Difference Between Centralized Edge AI and Distributed Edge AI}\label{table:summary}
\centering
\begin{tabular}{lll}
\rowcolor[HTML]{C0C0C0} 
\hline
\textbf{Parameters}       & \textbf{Centralized Edge Intelligence}                                            & \textbf{ZTP-enabled Distributed Edge AI}                               \\  \hline
Model            & Traditional supervised learning                 & Unsupervised and policy-based reinforcement learning                        \\
\rowcolor[HTML]{EFEFEF} 
Privacy          & No privacy for handling users data                  & Support privacy and security on data handling                 \\
Training Time    & Training on large data exponentially increases time & Training on local edge devices helps to optimize time         \\
\rowcolor[HTML]{EFEFEF} 
Heterogeneity    & Low                                                 & High                                                          \\
Scalability      & Not Scalable                                       & Highly scalable                                              \\
\rowcolor[HTML]{EFEFEF} 
Applications     & Traffic monitoring, data storage and analysis       & keystroke prediction, smart city, autonomous vehicle          \\
Computation Cost & incur high cost over the edge network               & As the model only shares learnable parameters, cost decreases \\
\rowcolor[HTML]{EFEFEF} 
Performance &
  \begin{tabular}[c]{@{}l@{}}Due to centralized architecture, edge AI \\ suffers from low accuracy\end{tabular} &
  \begin{tabular}[c]{@{}l@{}}As the model share knowledge, performance of the \\ network increases gradually\end{tabular} \\
Automation       & Medium                                                 & High        \\
\hline
\end{tabular}
\end{table*}

\subsection{Edge Resource Federation}
Standard edge computing and cloud computing models, delivering services to end-users, suffer from inflated resource mismanagement. There is a need for a unique and collective service provisioning strategy, where overcrowded edge devices interoperably communicate with nearby underloaded edge devices or cloud servers and share the excessive workload. 

Edge federation, also known as edge resource federation, is a combined resource provisioning strategy for edge networks. Edge federation manages the resources of the different edge devices offered by service providers and brings the edge resources into one platform \cite{cao2020edge}. In essence, edge federation aims for low latency, scalability, and cost-effectiveness by seamlessly integrating edge-to-edge and edge-to-cloud resources into one platform. Network Function Virtualization (NFV), Software-Defined Networking (SDN), containerization and container orchestration, as well as multi-access edge computing (MEC) are anticipated as critical enablers for automated edge resource federation. 

The edge federation model has two key advantages. First, it has the capability to gather edge services under one platform and handle dynamic service requests coming from different users while optimizing network resources and service delay. Second, edge federation combines different edge infrastructures and resources offered by different service providers by optimizing service deployment costs. As expected, combining DAI and edge federation techniques can introduce new responsive service assistance models, which could be a win-win solution for edge infrastructure providers, edge service providers, and end-users. Overall, we can summarize some of the benefits as follows:
\begin{itemize}
\item Reliable interconnection between edge and cloud.
\item Moving computing resources to the network edge.
\item Consistent user satisfaction ratio.
\item Building an edge hierarchy model.
\item Easy knowledge sharing among user devices.
\end{itemize}

\subsection{Distributed Intelligence}
In contrast to cloud AI, centralized edge AI trains ML models in nearby suitable computing devices and then deploys the models across distributed end devices, endowing the devices with local decision-making strategies~\cite{pujol2023edge}. However, centralized edge AI faces a number of challenges, including a lack of coordination among edge devices, a lack of global knowledge, and limited scope for edge federation. The present edge networks must be updated to use distributed intelligence, where edge devices can communicate and share end-device data models~\cite{peltonen2022many}. 

Distributed AI can solve complex understanding, learning, and decision-making problems by modeling them as multi-agent systems. The agents, or edge nodes in the DAI network, can operate independently and communicate asynchronously to combine partial solutions. Owing to the large data scale, DAI systems are resilient, flexible, and by definition, loosely connected. In contrast to monolithic or centralized AI systems, which have tightly connected and geographically close processing nodes, DAI systems do not require all relevant data to be gathered in a single location. Instead, many DAI systems work with small subsets of data, making them easy to employ. In Table~\ref{table:summary}, we have briefly presented the advantages of incorporating DAI in edge networks compared to traditional AI.

One of the most critical challenges in distributed edge computing is data gravity. Data gravity refers to the capability of a rich source of data to attract applications and services. Edge networks can be considered such rich data sources, with ZTP attracting users for edge services and applications while ensuring high throughput and optimized latency. 

Data gravity poses two fundamental issues. First, end-users place tremendous strain on the edge servers to manage all the generated and processed data, resulting in high processing costs for data analysis and training. Data gravity is solved by not collecting all data from end devices. Instead, only essential training data should be collected without noise in it.  

Another issue is the heterogeneity of edge devices. Edge devices are generally made by various infrastructure providers, and services have varying requirements. As a result, a model trained on an edge server will likely not fit all the other edge devices, making it always challenging for distributed edge networks. Therefore, the ZTP considers all such complex network- and system-based challenges into one frame and solves them using the DAI. Further, this framework has the capabilities to bring DAI to innumerable edge devices and allow it to scale across a wide variety of applications. Overall, we can summarize some of the key benefits as follows: 
\begin{itemize}
\item Improve the decision-making capabilities of local devices. 
\item Increase user data security and privacy.
\item Reduce data transmission costs to remote servers. 
\item Continuously update model and knowledge.   
\item Allow training with small and heterogeneous user data.
\end{itemize}

\textcolor{black}{In the context of Industry 4.0, implementing a smart factory highlights the benefits of utilizing DAI for ZTP. Within this particular environment, the edge devices situated within the factory exhibit AI capabilities that enable them to process sensor data in real time~\cite{hazra2021comprehensive}. For example, when a machine sensor detects a possible issue, edge AI promptly recognizes it, implements corrective measures, and reduces the delay in critical decision-making. On the other hand, a centralized or cloud-based AI system necessitates data transmission to a distant location for analysis, potentially causing unfavourable delays and operational hazards. This example demonstrates the considerable enhancement of ZTP through the implementation of DAI at the edge, with a special focus on its impact in the context of Industry 4.0. This approach effectively improves production efficiency and reduces downtime by facilitating real-time, localized, and informed decision-making.}

\subsection{Zero-touch Provisioning}
There is a trend toward ever more on-demand offering of storage and resource management capabilities. With the increasing number of resources being managed,  delivering and managing dynamic user service requests becomes ever more complex. To overcome this complexity, ETSI offers the idea of zero-touch network provisioning as a new breed of network management functionality, seeking to integrate network functionality, cutting-edge communication technologies (eMBB, URLLC, and mMTC), as well as automatically carrying out edge computing processes. 

Distributed AI is expected to be a key facilitator of self-learning capabilities, leading to lower operating costs, quicker time-to-value processes, and a smaller chance of human errors. Although there is a rising desire to use DAI in a ZTP network, there may also be limitations and risks associated with doing so. The abilities of ZTP networks are specified on fully combined self-3s life cycle functions (\textit{i.e.,} self-fulfilling,  self-serving, and self-assuring) to automatically satisfy and respond to customer resource demands. However, to implement this in real-time, we need to take advantage of network controllers and advanced communication technologies such as 5G or 6G. 

\begin{figure}[!t]
    \centering
    \includegraphics[width=0.5\textwidth]{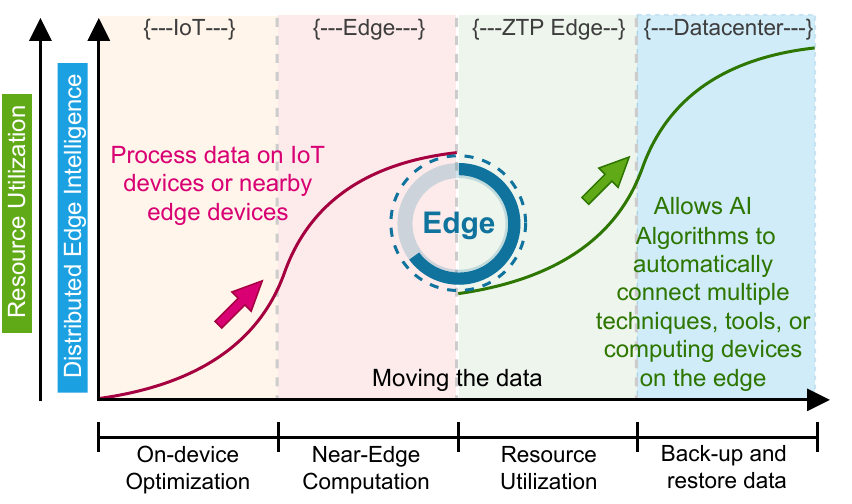}
    \caption{Computation strategy in zero-touch edge networks.}
    \label{fig3}
\end{figure}

Inspired by existing cloud service models, such as security-as-a-service, database-as-a-service, etc., ZTP can also be provided to end-users as a service model. Under this umbrella, computation, communication, and ephemeral storage can be provided to the end user or other IoT vendors. ZTP providers should be able to personalize the resources the customer would like to avail based on their needs. For instance, customers could manage the entire life cycles of edge applications on their IoT devices, including application deployment, configuration, starting, and stopping. This requires managing the computational and storage resources on each device, as well as the communication resources for message exchange and data flow among edge application components. 

In addition, multiple vendors could install specific IoT devices (with additional computation and storage resources) using ZTP technology. This transformation enables computation execution near the data sources, as presented in Fig.~\ref{fig3}\footnote{Inspired from Distributed Artificial Intelligence at the Edge and beyond, \url{https://engineering.cmu.edu/accelerator/news/2021/03/03-ai-fusion.html}}. In such a multi-vendor edge infrastructure, idle resources on individual edge devices can be rented out to other vendors as a service. 
Overall, we can summarize some of the benefits as follows. 
\begin{itemize}
    \item 100\% full automation of network devices.           
    \item Shorter time for execution on remote servers.
    \item Reduce the chances of human errors.    
    \item Easy to fix and auto upgrade technical programs.
    \item Easy upgrade of hardware equipment.
\end{itemize}

\section{Potential Challenges and Future Directions}\label{section3}
This section provides a list of challenges and possible future research directions for implementing ZTP and the usability of DAI in Edge networks. 
\subsection{Challenges}
The ultimate goal of ZTP is to add convenience to edge network management, limiting human intervention. However, using ZTP in edge networks does entail challenges. 
\begin{itemize}
    \item \textit{Cascading Failures: } With cascading failures, a low-level failure may lead to failures also on higher levels. ZTP has no mechanism to control such cascades. Instead, ZTP may report a single problem as multiple, making failure analysis more difficult.  
    \item \textit{Anomaly Detection} The ZTP model does not support, maintain, or automate service lifecycles in the entire computing continuum. Moreover, while monitoring services, ZTP has no mechanisms for causing alarms on individual, anomalous activities (e.g., faulty nodes), and few for responding to them. 
    \item \textit{\color{black}Data Heterogeneity:}The concept of distributed computing continuum data heterogeneity spans a wide range of data types, sources, and spatiotemporal properties. The significance of this lies in its ability to offer extensive perspectives, tailor-made solutions and facilitate data integration for a more nuanced understanding~\cite{hazra2021comprehensive}. Nevertheless, certain issues need to be addressed in distributed networks. These challenges encompass data integration, quality, and scalability complications in broad and diverse environments.

    \item \textit{Limits Orchestration: } ZTP can automate small tasks and initial setups such as activating licenses, running containerized apps, bootstrapping virtual machines, and even updating device firmware. However, current ZTP implementations lack mechanisms for automating processes and workloads. We consider it a challenge because manual work degrades the value of zero touch. 
    \item \textit{Security: } There is a vast number of connected devices with continuous services in the computing continuum. Maintaining security on autonomous systems running on those devices with no human intervention is more challenging. DAI may help to design efficient and dynamic mechanisms across the continuum to detect unforeseen threats or vulnerabilities.
\end{itemize}
\subsection{Research Directions}
This section fills this gap by providing possible open challenges for further research. 
\begin{itemize}
\item \textit{Light-weight AI/ML}:
Resource-constrained end devices and edge nodes need low latency. Light-weight AI/ML algorithms minimize both resource usage as well as the time spent computing without affecting the prediction accuracy. ML model compression, which reduces the amount of redundant data in the models, is one way of achieving lightweight ML models. However, novel methods for lightweight AI/ML algorithms in ZTP are needed to further increase energy efficiency in edge networks. 
\item \textit{Semantic Interoperability}:
The computing continuum interconnects a set of devices that are heterogeneous in terms of, e.g., technologies, device standards, data formats, etc. This lack of interoperability limits the utility of ZTP in the computing continuum. It is thus necessary to bridge the gap between the ZTP and the computing continuum by developing intelligent interoperable protocols.

\item \textit{Privacy}:
IoT, cloud, data centers, gateways, etc., are all generating and exchanging massive volumes of sensitive data. The privacy of these data must be ensured while designing the ZTP for the computing continuum, as ZTP precludes human intervention. 

\item \textit{Low-latency: }
A number of time-critical use case scenarios such as medical, industry, smart city, etc, require rapid decisions. Designing low-latency mechanisms in ZTP is thus essential for the computing continuum. Future research can focus on developing techniques through intelligent agents that can prioritize time-critical requests and process them autonomously.

\item {\color{black}\textit{ZTP for Intelligent Protocols:}  There is an ever-growing number of computing devices in the computing continuum, and a vast number of data transmissions between them, so developing adaptive and intelligent data protocols is challenging. In this context, ZTP can help fault diagnosis and autonomous decision making mechanisms in these protocols. In particular, broker-based publish/subscribe communication patterns may benefit more from ZTP and distributed AI, which may increase their adaptability and efficiency. There is huge scope for research into making existing data protocols intelligent with the help of ZTP. }

\item \textcolor{black}{\textit{Explainability: }
ZTP will autonomously select configuration states for large distributed systems, which will determine their behavior. In that regard, is crucial to develop \textit{sidecar} tools able to explain why that specific configuration was selected. To do that, causality is emerging as a candidate technology to provide explainability to self-adaptive systems.}

\item {\color{black}\textit{Generative AI for ZTP:} In general, AI or ML techniques can predict issues by analyzing data. However, all these predictions are likely to be expected. In view of the computing continuum's complexity constraints, there is a possibility for unpredictable issues in the future. It is possible to identify or solve unpredictable issues within the systems using recent advances in large language models and generative AI technology. It remains challenging to identify potential computing nodes to perform generative AI in the computing continuum. Also, tracing accuracy of generative AI decisions on-the-fly is another challenging. Further research on the use of generative AI for ZTP must provide additional benefits to the computing continuum as a whole. }

\end{itemize}

\section{Conclusion}\label{section4}
In this article, we showed the benefits of combining distributed AI and Zero-touch Provisioning in the device-edge-cloud computing continuum. We discussed the pivotal role that distributed AI approaches maintain in creating ZTP. Moreover, we emphasized the constraints and challenges that may impede the integration of distributed AI in edge-enabled ZTP networks. We also shed light on several potential research solutions for establishing an intelligent and autonomous edge environment in light of the specified research challenges.

\section*{Acknowledgment}
This research is partially supported by the following projects: (i) the Academy of Finland through the 6G Flagship program (Grant 318927); (ii) the European Commission and select member countries through the ECSEL JU FRACTAL project (Grant 877056); (iii) Business Finland through the Neural pub/sub research project (diary number 8754/31/2022).

\bibliographystyle{IEEEtran}
	\bibliography{ref}
\vskip -1\baselineskip plus -1fil
\begin{IEEEbiographynophoto}{Abhishek Hazra (M'18, M'22)}~Abhishek Hazra is an Assistant Professor in the Department of Computer Science and Engineering at the Indian Institute of Information Technology, Sri City. He has also worked as a Post-doctoral Research Fellow at the Communications and Networks Lab, Department of Electrical and Computer Engineering, NUS Singapore. He earned his Ph.D. from IIT(ISM) Dhanbad, India, and holds a master's degree in Computer Science and Engineering from NIT Manipur, India, as well as a bachelor's degree from NIT Agartala, India. Currently, he serves as an Editor for Computer Communication and Physical Communications. Abhishek Hazra has authored and co-authored numerous national and international journal and conference articles. His research interests include IoT, Fog/Edge Computing, 6G, Machine Learning, and Industry 4.0/5.0. (Contact him at \texttt{a.hazra@ieee.org})

\end{IEEEbiographynophoto}\vskip -1\baselineskip plus -1fil
\begin{IEEEbiographynophoto}{Andrea~Morichetta~(M'22)} 
is a university assistant with the Distributed Systems Group, TU Wien, 1040, Vienna, Austria.
His research interests include the intersection of complex distributed systems and machine learning, network monitoring, and security. Morichetta received his Ph.D. degree in machine learning and big data approaches for automatic Internet monitoring from the Telecommunication Network Group, Politecnico di Torino, Turin, Italy. He is a Member of IEEE. Contact him at \texttt{a.morichetta@dsg.tuwien.ac.at}.
\end{IEEEbiographynophoto}\vskip -1\baselineskip plus -1fil

\begin{IEEEbiographynophoto}{Ilir Murturi~(M'19)}~is currently a Postdoctoral Researcher with the Distributed Systems Group, TU Wien, Vienna. He received his M.Sc. degree in Computer Engineering from the University of Prishtina, Kosova. He received his Ph.D. from the Faculty of Informatics, Technical University of Vienna (TU Wien) in 2022. His current research interests include the Internet of Things, Edge Computing, EdgeAI, self-adaptive and cyber-physical systems. He is an IEEE Member and serves as Guest Editor for the International Journal of Network Management, Wiley. (Contact him at \texttt{i.murturi@dsg.tuwien.ac.at}).
\end{IEEEbiographynophoto}\vskip -1\baselineskip plus -1fil

\begin{IEEEbiographynophoto}{Lauri Lovén (SM'19)},~D.Sc.(Tech), is a postdoctoral researcher and the coordinator of the Distributed Intelligence strategic research area in the 6G Flagship research program, at the Center for Ubiquitous Computing (UBICOMP), University of Oulu, in Finland. He received his D.Sc. at the University of Oulu in 2021, was with the Distributed Systems Group, TU Wien in 2022, and visited the Integrated Systems Laboratory at the ETH Zürich in 2023. His current research concentrates on edge intelligence, and on the orchestration of resources as well as distributed learning and decision-making in the computing continuum. He has co-authored 2 patents and ca. 50 research articles. (Contact him at \texttt{lauri.loven@oulu.fi}).
\end{IEEEbiographynophoto}\vskip -1\baselineskip plus -1fil

\begin{IEEEbiographynophoto}{Chinmaya~Kumar~Dehury~(M'16)}~is Assistant Professor of Distributed Systems, member of Mobile \& Cloud Lab in the Institute of Computer Science, University of Tartu, Estonia. He received his Ph.D. Degree in the Department of Computer Science and Information Engineering, Chang Gung University, Taiwan. His research interests include scheduling, resource management and fault tolerance problems of Cloud, fog Computing, and clustered edge computing, IoT, and data management frameworks. He is a member of IEEE and ACM India. He is also serving as a PC member of several conferences and reviewer to several journals and conferences, such as IEEE TPDS, IEEE JSAC, IEEE TCC, IEEE TNNLS, Wiley Software: Practice and Experience, etc. (Contact him at \texttt{chinmaya.dehury@ut.ee})
\end{IEEEbiographynophoto}\vskip -1\baselineskip plus -1fil
\begin{IEEEbiographynophoto}{Víctor Casamayor Pujol~(M'22)}
is a project assistant (Postdoc) in the Distributed Systems Group at TU Wien. In 2020 he obtained his PhD in ICT by Universitat Pompeu Fabra in Barcelona, Spain. He has a master in Intelligent Interactive Systems (MIIS) by UPF in Barcelona, Spain and a specialized master in space systems engineering (TAS-Astro) by ISAE in Toulouse, France. He has also worked in space propulsion at the CNES in Paris, France. His research interests revolve around self-adaptive methodologies for computing continuum systems, including SLO-based definitions, causal and machine learning inference, and robotics. (Contact him at \texttt{v.casamayor@dsg.tuwien.ac.at})
\end{IEEEbiographynophoto}\vskip -1\baselineskip plus -1fil

\begin{IEEEbiographynophoto}{Praveen Kumar Donta (SM'22)}~is a Postdoctoral researcher in the Distributed Systems Group, TU Wien, Austria since July 2021. He received his PhD from the Department of Computer Science and Engineering at IIT (ISM), Dhanbad, India in June 2021. He was a visiting PhD student for six months during his PhD at Mobile \& Cloud Lab, University of Tartu, Estonia. He received his M.Tech and B.Tech from JNT University Anantapur, India in 2014, and 2012. He is a member of IEEE Computer and Communications Societies. He is serving as Editor for Computer Communications, Measurement, and Measurement: Sensors, Elsevier, ETT, Wiley, and PLOS One. His current research includes Learning-driven distributed computing continuum systems. (Contact him at \texttt{p.donta@dsg.tuwien.ac.at}).
\end{IEEEbiographynophoto}\vskip -1\baselineskip plus -1fil

\begin{IEEEbiographynophoto}{Schahram~Dustdar~(F'16)}~is Full Professor of Computer Science heading the Research Division of Distributed Systems at the TU Wien, Austria. He is Editor-in-Chief of Computing (Springer). He is an Associate Editor of IEEE Transactions on Services Computing, IEEE Transactions on Cloud Computing, ACM Transactions on the Web, and ACM Transactions on Internet Technology, as well as on the editorial board of IEEE Internet Computing and IEEE Computer. Prof. Dustdar is Recipient of the ACM Distinguished Scientist Award (2009), the ACM Distinguished Speaker ward (2021), the IBM Faculty Award (2012), an Elected Member of the Academia Europaea: The Academy of Europe, where he is Chairman of the Informatics Section, as well as an IEEE Fellow. In 2021 Dustdar was elected  President for Asia-Pacific Artificial Intelligence Association (AAIA). (contact him at \texttt{dustdar@dsg.tuwien.ac.at})
\end{IEEEbiographynophoto}

\end{document}